\pdfoutput=1

\documentclass[11pt]{article}

\usepackage{acl}

\usepackage{times}
\usepackage{latexsym}

\usepackage[T1]{fontenc}

\usepackage[utf8]{inputenc}

\usepackage{microtype}

\usepackage{inconsolata}

\usepackage{todonotes}
\usepackage{booktabs, supertabular, multirow, longtable, tabularx}
\usepackage{caption}
\usepackage{subcaption}
\usepackage{graphicx}
\usepackage{amssymb}
\usepackage{relsize}
\usepackage{siunitx}
\usepackage[inline]{enumitem}
\usepackage{amsmath}
\usepackage{tablefootnote}
\usepackage[inline]{enumitem}
\usepackage{fixltx2e}
\usepackage{etoolbox}
\usepackage{makecell}
\usepackage{tabularx}
\usepackage{xcolor}
\usepackage{nicefrac}
\usepackage{dblfloatfix}

\definecolor{subblue}{HTML}{1f77b4}
\definecolor{objorange}{HTML}{ff7f0e}
\definecolor{customred}{HTML}{d62728}
\definecolor{customgreen}{HTML}{2ca02c}  

\title{BEAR: A Unified Framework for Evaluating Relational Knowledge in Causal and Masked Language Models}

\author{
    Jacek Wiland*\\
    \And
    Max Ploner* \vspace*{3mm}\\
    Humboldt Universität zu Berlin \\
    Science Of Intelligence\\
    \{\texttt{jacek.wiland}, \texttt{max.ploner}, \texttt{alan.akbik}\}\texttt{@hu-berlin.de}
    \And
    Alan Akbik
    \\
}

\begin{document}
\pagestyle{plain}

\maketitle
\begingroup\def\thefootnote{*}\footnotetext{Equal contribution}\endgroup

\begin{abstract}
  Knowledge probing assesses to which degree a language model (LM) has successfully learned relational knowledge during pre-training. Probing is an inexpensive way to compare LMs of different sizes and training configurations. However, previous approaches rely on the objective function used in pre-training LMs and are thus applicable only to masked or causal LMs. As a result, comparing different types of LMs becomes impossible. To address this, we propose an approach that uses an LM's inherent ability to estimate the log-likelihood of any given textual statement. We carefully design an evaluation dataset of 7,731 instances (40,916 in a larger variant) from which we produce alternative statements for each relational fact, one of which is correct. We then evaluate whether an LM correctly assigns the highest log-likelihood to the correct statement. Our experimental evaluation of 22 common LMs shows that our proposed framework, BEAR, can effectively probe for knowledge across different LM types. We release the BEAR datasets and an open-source framework that implements the probing approach to the research community to facilitate the evaluation and development of LMs. 
\end{abstract}

\section{Introduction}

Pre-trained language models (LMs) are the backbone of current state-of-the-art NLP approaches. A key property is the syntactic and semantic knowledge stored in their internal parameters, allowing them to generalize beyond given training data when fine-tuning for a specific downstream NLP task. Due to their importance and the large number of proposed LMs, prior work has sought to improve the ability to measure the amount of factual knowledge encoded in LMs, thereby facilitating the comparison of different LMs~\cite{petroniLanguageModelsKnowledge2019,poernerEBERTEfficientYetEffectiveEntity2020,caoKnowledgeableEducatedGuess2021,kaloKAMELKnowledgeAnalysis2022}. 

The LAMA probe~\cite{petroniLanguageModelsKnowledge2019} is the seminal work in studying commonsense and relational knowledge in LMs and is widely used for inexpensive evaluation and model comparison (see~\citet{youssefGiveMeFacts2023} and \citet{caoLifeCycleKnowledge2023} for an overview). 
The idea is to use relational knowledge from an existing knowledge base (KB) and create cloze-style statements for an LM to fill in. 

\begin{figure}[t]
  \centering
  \begin{subfigure}{\linewidth}
    \includegraphics[width=\linewidth, clip, trim={0cm 2.9cm 0cm 0.1cm}]{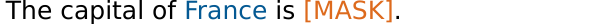}
    \vspace*{-3mm}
    \caption{LAMA probe: Single-subtoken mask prediction}
    \label{fig:LAMA}
  \end{subfigure}

  \vspace*{3mm}
  
  \begin{subfigure}{\linewidth}
    \includegraphics[width=\linewidth, clip, trim={0cm 0cm 0cm 1.2cm}]{illustrations/LAMA_vs_BEAR.pdf}
    \vspace*{-3mm}
    \caption{BEAR probe: Rank answer options of arbitrary length}
    \label{fig:BEAR}
  \end{subfigure}

  \caption{
    Comparison of the LAMA and BEAR probes.
    Both probes query LMs given a template (here in black), the subject of the relation (blue), and the object (orange). LAMA masks the object and predicts a single token as the answer. In BEAR, we create separate textual statements for a set of potential answers and select the statement with the highest (pseudo) log-likelihood as assigned by the LM. This method allows us to include multi-token answers and evaluate causal and masked LMs. }
  \label{fig:lama_vs_bear}
  \vspace*{-0.3cm}
\end{figure}

For instance, the entities ``France'' and ``Paris'' may be connected through the \textsc{has-capital} relation in a given KB, indicating that Paris is the capital of France. From this, LAMA constructs the sentence ``The capital of France is [MASK]'' and evaluates whether an LM predicts the correct token to complete this factual sentence. LAMA, therefore, effectively reuses the masked language modeling objective of the bidirectional family of LMs~\cite{devlinBERTPretrainingDeep2019b} to probe for knowledge. This example is shown in  Figure~\ref{fig:LAMA}. 

\paragraph{Limitations of LAMA.} While LAMA offers a straightforward approach to probing, it also has significant limitations.

First, LAMA requires the correct answer to be part of the evaluated LM's subtoken vocabulary, restricting the relational knowledge that can be tested to single-subtoken answers (such as ``Paris'' in Figure~\ref{fig:LAMA}). Thus, it cannot test for relational facts with long or rare answers (as shown in Figure~\ref{fig:BEAR}). 

Second, and most importantly, LAMA relies on the masked language modeling objective.
This makes LAMA inapplicable for LMs trained with other objectives. Therefore, it excludes causal LMs such as the GPT-family of models~\cite{radfordLanguageModelsAre2019}. To the best of our knowledge, no factual knowledge probe currently exists that applies to both masked and causal LMs. 

Third, various prior works have noted limitations of the relational data used in the LAMA probe, such as (1) a heavily skewed answer space, favoring some answers over all others~\cite{jiangHowCanWe2020, zhongFactualProbingMASK2021,caoKnowledgeableEducatedGuess2021}, (2) overly revealing entity names~\cite{poernerEBERTEfficientYetEffectiveEntity2020}, (3) and issues involving knowledge with multiple correct answers, causing correct answers to be counted as errors~\cite{kaloKAMELKnowledgeAnalysis2022}.    

These limitations of both the probing approach and the data impair LAMA's ability to measure and compare the relational knowledge of different LMs accurately.
 
\paragraph{Contributions.} To address these issues, we propose BEAR, a unified knowledge probe for both causal and masked LMs. Instead of casting the evaluation as a token prediction problem over the entire vocabulary of an LM, we present a set of answer options for each relation instance, create a textual statement for each option, and leverage the inherent ability of each LM to assign a log-likelihood score to statements, thereby ranking these options. See Figure~\ref{fig:BEAR} for an illustration. 

We argue that this approach has numerous benefits in that it (1) allows us to evaluate both masked and causal LMs, (2) imposes no restrictions on the answer space, (3) allows us to design a new evaluation dataset that addresses a range of issues such as answer skews and multiple correct answers noted in prior work. 
In more detail, our contributions are: 

\begin{enumerate}
  \item We present an analysis of the weaknesses of the LAMA probe and follow-up works to derive desiderata for the BEAR probe (see Section~\ref{sec:background}). 

  \item We propose to query knowledge as a multiple-choice selection problem in which an LM estimates the (pseudo) log-likelihood of a given answer template with each choice filled in (see Section~\ref{sec:bear_probe}). 
  
  \item We construct a novel evaluation dataset that reflects the desiderata identified in our analysis (see Section~\ref{sec:bear_data}). 
  
  \item We use BEAR to evaluate a range of common masked and causal LMs (see Section~\ref{sec:experiments}). 
  
\end{enumerate}

To enable the community to employ the proposed probing method and dataset, we publicly release\footnote{The library, the probing dataset, as well experimental artifacts can be retrieved via the following URL:\\ \url{https://lm-pub-quiz.github.io/}} the evaluation framework \textit{lm-pub-quiz} \citep[based on the \textit{minicons},][]{misra2022minicons} as well as the dataset \textit{BEAR}\footnote{\textbf{B}enchmark for \textbf{E}valuating \textbf{A}ssociative \textbf{R}easoning), \emph{CC BY-SA} license.}. 

\section{LAMA and Follow-Up Work}
\label{sec:background}

We discuss the technical details of the LAMA probe first, followed by an analysis of its weaknesses. 

\paragraph{LAMA evaluation data.} The LAMA benchmark was originally composed of four separate datasets named after their respective sources: SQuAD \citep{rajpurkarSQuAD1000002016c}, GoogleRE\footnote{\url{https://code.google.com/archive/p/relation-extraction-corpus/}}, ConceptNET \citep{speerRepresentingGeneralRelational2012}, and T-REx \citep{elsaharTRExLargeScale2018}. However, subsequent research mainly focused exclusively on the T-REx subset. 
Its knowledge base comprises 41  \textit{relations} derived from Wikidata. Each relation contains at most 1,000 \textit{relation instances} in the form of subject-relation-object triples  $\langle s, r, o \rangle$ where $s$ is the subject (e.g., ``France''), $r$ the relation (e.g., \textsc{has-capital}), and $o$ an object (e.g.~``Paris'').

There are three types of relations in LAMA: 1-1 (one-to-one, e.g., \textsc{has-capital}), N-1 (many-to-one, e.g., \textsc{has-language}), and N-M (many-to-many, e.g., \textsc{shares-border-with}). Relations of the N-1 type allow multiple subjects to relate to one object, while the latter permits many subjects to be associated with numerous objects. 

\paragraph{Relation identifiers.} All relations are linked to a corresponding relation in Wikidata and thus have unique IDs. For instance, the \textsc{capital-of} relation in LAMA corresponds to Wikidata relation P1376 (see Table~\ref{tab:multipletoken-ratio} for more examples). It facilitates comparison across different datasets since all follow-up works to LAMA, including BEAR, derive their relations from Wikidata. 

\paragraph{Templates.} Each relation in LAMA has a textual template with placeholders for subject and object. For \textsc{capital-of}, the template is ``[X] is the capital of [Y].'', where [X] is a placeholder for the subject, while [Y] is the placeholder for the object. At test time, the subject of a given relation is filled in the template, while the object is replaced by a [MASK]-token. This procedure results in a masked sentence (e.g., ``Paris is the capital of [MASK].'') for which the LM is tasked to predict the masked token.

\subsection{Issue 1: Single Subtoken Answers}

As noted by \citet{petroniLanguageModelsKnowledge2019}, LAMA is restricted to single-subtoken answers for factual knowledge queries. This limitation causes issues as LMs split most words into multiple subtokens, and most LMs differ in how they perform the splits. To illustrate, consider how the country name ``Togo'' is tokenized by different versions of BERT: the \texttt{bert-base-cased} model splits the word into two subtokens ([\texttt{To}, \texttt{\#\#go}]), whereas the \texttt{bert-base-uncased} variant preserves it as a single subtoken ([\texttt{togo}]). 

An analysis of 194 UN member country names is an excellent example of how such a restrictive condition affects the size of a hypothetical dataset. When restricting answer space to single tokens, 32\% and 27\% of available country names would have to be discarded for cased and uncased versions of BERT, respectively. Worse, the RoBERTa models~\cite{liuRoBERTaRobustlyOptimized2019} that uses a BPE-based tokenizer would split 88\% of all country names. Refer to Table~\ref{tab:multipletoken-ratio} for a list of how many LAMA instances need to be discarded when evaluating \texttt{xlm-roberta-base} \citep{conneauUnsupervisedCrosslingualRepresentation2020a} and  \texttt{roberta-base} models.

\paragraph{Comparison of different LMs.} Because the tokenizer bundled with each model differs, comparing various LMs becomes only possible if the models tokenize the answers in the same way. To address this, practitioners are currently reverting to using the intersection of single-token vocabularies derived from all LMs being compared. However, in practice, this further limits the scope of relational knowledge that can be included in the evaluation. 

\begin{table}
  \centering
  \small
  \resizebox{\columnwidth}{!}{%
  \begin{tabular}{cc cc}
    \toprule
    ID &  Relation & \texttt{xlm-roberta-base} &  \texttt{roberta-base} \\
    \midrule
    P30 &   \textsc{on-continent} &         74.46 \%&         80.21\% \\
    P31 &  \textsc{instance-of} &         28.85\% &         67.35\% \\
    P36 &   \textsc{has-capital} &         45.80\% &         89.76\% \\
    P37 &   \textsc{has-language} &         30.85\% &         45.13\% \\
    ... &    ... &        ...   &       ... \\
    P1303 &   \textsc{instrument} &   58.69\% &        100.00\% \\
    P1376 &   \textsc{capital-of} &    32.05\% &         81.62\% \\
    \midrule
    \multicolumn{2}{c}{\textbf{Mean}} & 31.73\% & 62.86\% \\
    \bottomrule
  \end{tabular}
  \vspace*{-0.2cm}
  }
  \caption{Ratio of discarded instances due to multi-token answers in \texttt{xlm-roberta-base} and \texttt{roberta-base}.}
  \label{tab:multipletoken-ratio}
  \vspace*{-0.2cm}
\end{table}

\paragraph{Prior work.} Various prior works address the issue of predicting multi-subtoken words for single [MASK] tokens \cite{ghazvininejadMaskPredictParallelDecoding2019, jiangXFACTRMultilingualFactual2020, kalinskySimpleEffectiveMultiToken2023, shenBlankLanguageModels2020, robinsonLeveragingLargeLanguage2023}. \citet{jiangXFACTRMultilingualFactual2020} provided a selection of algorithms to tackle predicting multi-token entities. However, they require a specification of further parameters, such as the number of subtokens to generate. \citet{kalinskySimpleEffectiveMultiToken2023} proposed generation approaches that either require additional training or the use of an external network, making them inapplicable to the purpose of evaluating knowledge contained in pre-trained weights through a zero-shot approach. 

\begin{figure}[b!]
  \centering
  \includegraphics[width=0.5\textwidth]{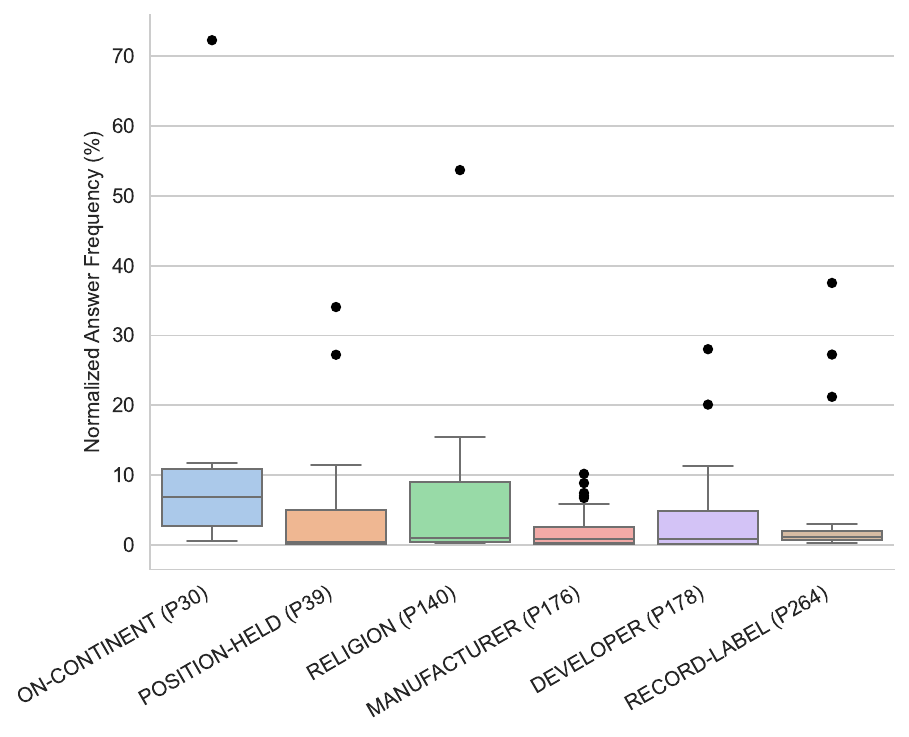}
  \vspace*{-0.8cm}
  \caption{The normalized answer frequency of selected relations in the LAMA probe. The outliers are marked with dots. In some relations, a majority class accounts for more than 50\% of all instances. }
  \label{fig:trex_ans_freq_dist}
\end{figure}

\begin{figure*}[t!]
  \vspace*{-0.4cm}
    \includegraphics[width=\linewidth]{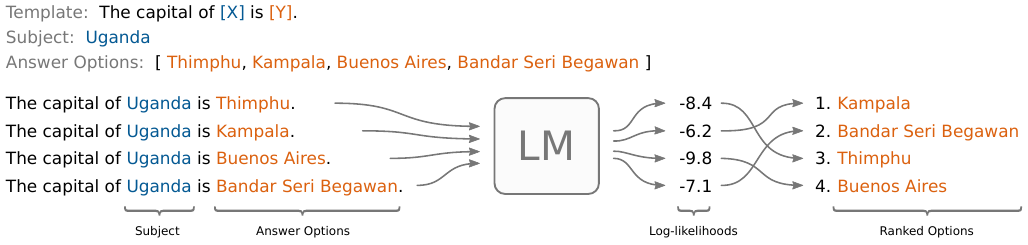}
    \caption{For each answer option, a sentence is passed to the LM (here using the template: ``The capital of [X] is [Y].'' and the subject ``Uganda''). The log-likelihood scores assigned by the LM are then used to rank the answer options. }\label{fig:bear_eval_illustration}
  \vspace*{-0.2cm}
\end{figure*}

\subsection{Issue 2: Multiple Correct Answers}

LAMA expects exactly one correct answer to each knowledge query and rates other factually correct answers as errors. To illustrate this, consider the query ``Germany shares a border with [MASK]'', to which LAMA expects the answer ``Poland''. All other correct answers, such as ``Denmark'' are marked as incorrect. This issue affects all N-M relations in LAMA. 

\paragraph{Prior work.}
KAMEL~\cite{kaloKAMELKnowledgeAnalysis2022} address this by allowing the LM to generate an arbitrary number of answers using a template instructed via few-shot prompting, experimenting with ranges of 1-10 answers per instance. Subsequently, they evaluate the predictions using standard measures of precision and recall. However, their approach relies on the text generation ability of causal LMs and thus cannot be applied to masked LMs.

\subsection{Issue 3: Imbalanced Answer Distribution}

The relations in T-REx have a highly unbalanced answer distribution (except the 1-1 relations), and in certain relationships, over half of the instances belong to the predominant class (see Figure~\ref{fig:trex_ans_freq_dist}). \citet{zhongFactualProbingMASK2021} noted that a model that always chooses the majority class outperforms some state-of-the-art LMs on selected relations.

To illustrate, consider the T-REx's \textsc{on-continent} relation, which connects a location to the continent in which it is situated. Counter-intuitively, the majority class in this relation is ``Antarctica'', accounting for $72\%$ of all instances. 

\paragraph{Prior work.} 
To account for this imbalance, \citet{caoKnowledgeableEducatedGuess2021} created a balanced version of the LAMA probe called WikiUNI. It contains the same relations as T-REx but has a uniform answer distribution and was constructed to have the same number of subjects for every object. However, their dataset samples a highly skewed number of instances per relation, with 7 relations (out of 41) accounting for over 50\% of all instances.

\subsection{Issue 4: Rare Wikidata Entries}
\label{sec:rare-wiki-entries}

The above-mentioned example of ``Antarctica'' being an answer to over $72\%$ of all instances in the  \textsc{on-continent} relation also points to another problem: An artifact of randomly sampling Wikidata for relation instances is that rare Wikidata entries are overrepresented. For example, \textsc{on-continent} has a large number of small islands as subjects (e.g., ``Umber Island'' and ``Brooklyn Island'', both close to the Antarctic continent), many of which are unlikely to occur in a corpus outside of an encyclopedia like Wikipedia. 
We believe that the LAMA dataset unfairly favors LMs trained using Wikipedia. However, to the best of our knowledge, no prior work has addressed this issue.

\subsection{Issue 5: Evaluation of Causal LMs}

LAMA relies on the capability of masked language models to fill in masked tokens, making it unsuitable for causal LMs.

\textbf{Prior work.}
To address this, \citet{kaloKAMELKnowledgeAnalysis2022} proposed the KAMEL probe. Factual knowledge is probed by virtue of question statements for which the response is auto-regressively generated using the causal LM. To guide the generation approach, they prepend $k$ few-shot examples into the prompt that present how the correctly formatted answer should look.  
However, since this approach relies on the language modeling objective of causal LMs, KAMEL does not apply to masked LMs. 

\subsection*{Further Related Work}

Further related work has attempted to mitigate the sensitivity of the  evaluation on the timing of pre-training (and hence the corpus used) 
\citep{dhingraTimeAwareLanguageModels2022a, mallenWhenNotTrust2023a, onoeEntityClozeDate2022a}.
Nonetheless, as these studies do not directly address the concerns outlined in the previous subsections, we will not discuss them in detail. Interested readers are referred to the cited papers for more information. 

\section{BEAR Probe}
\label{sec:bear_probe}

We base our evaluation on using an LM's inherent ability to estimate the log-likelihood of a given sentence. Our main idea is to restrict the space of possible answers for each relation instance and create a set of options that are then ranked by their log-likelihood values.

\subsection{Ranking Options using Log-Likelihood}

Our approach requires a dataset of $\langle s, r, o \rangle$ relation instances, where for each relation $r$ there exists (at least) one template $t$ and a set of answer options $a_i$ with $i\in\{1, ..., k\}$ that includes the correct answer. 

\paragraph{Creating options to rank.} For each relation instance, we create $k$ natural language statements using the template, instance's subject $s$, and each of the possible relation's objects $a_i$ as parts of a textual statement. 

Figure~\ref{fig:bear_eval_illustration} illustrates this process for the example relation instance $\langle$``Uganda'', \textsc{has-capital}, ``Kampala''$\rangle$ and the template ``The capital of [X] is~[Y]''. This example's set of potential answers is [``Kampala'', ``Thimphu'', ``Buenos Aires'', ``Bandar Seri Begawan'']. For each potential answer, we create a separate statement.

\paragraph{Predicting log-likelihood.} For each generated statement, we predict the log-likelihood score $\log \hat{p}(a | t)$. As the template is the same for each of the answer options (i.e., $\hat p(t)$ is constant) and we are only interested in ranking them, it is sufficient to compute the log-likelihood for the entire sentence:
$$
\log \hat p(a_i | t) = \log \hat p(a_i, t) - \log \hat p(t)
$$

Since causal LMs are trained to predict a log-likelihood of each token given the previous context, the log-likelihood of a sentence is simply the sum of the log-likelihoods of each token.

\paragraph{Log-likelihood in masked LMs.}  
A sentence-level log-likelihood is not clearly defined for an LM trained using the masked language modeling objective.
However, \citet{salazarMaskedLanguageModel2020} and \citet{kaufBetterWayMasked2023b} offer two variants of how to retrieve a pseudo log-likelihood score for a given text.
Both approaches use multiple forward passes. \citet{salazarMaskedLanguageModel2020} simply mask each token once while keeping the remaining context unmasked. 
\citet{kaufBetterWayMasked2023b} improve on this by additionally masking all tokens right to the current token belonging to the same word. 
This approach fixes the issue of assigning disproportionate likelihoods to multi-token words. We use the latter in our approach.

\paragraph{Ranking the results.} Finally, the statements are ranked by their log-likelihood scores. This is illustrated in Figure~\ref{fig:bear_eval_illustration} (right-hand side).

\subsection{Evaluation Metric}

To evaluate the amount of knowledge encoded in each model, we score whether the top-ranked statement is the correct answer for each relation instance.
Previous work \citep[such as][]{petroniLanguageModelsKnowledge2019} additionally considered answers with higher ranks.
The mean precision $P@k$ (for a given rank $k$) is commonly reported.

However, given that we evaluate on a constrained answers space, we believe the first rank to be sufficient.
In the case of $k=1$, this is identical to the accuracy metric.
We report the average over all relation instances in our evaluation data as the BEAR score.

\section{BEAR Dataset}
\label{sec:bear_data}

Our proposed probing approach requires a dataset with a restricted answer space. Following the analysis in Section~\ref{sec:background} we additionally desire 
\begin{enumerate*}[label={(\arabic*)}]
    \item the answer space to be balanced, 
    \item a single correct answer per relation instance,
    \item a balanced number of instances per relation, and
    \item a focus on knowledge that could reasonably be found in corpora other than Wikipedia.
\end{enumerate*}

\subsection{Selecting Relations}

We use the 234 relations of KAMEL as a starting point and manually remove two-thirds of these.
This curation process was conducted independently by two researchers (authors of this paper), and disagreeing judgments were discussed in detail to reach a final decision. 
The most common reasons for excluding a relation were:

\begin{itemize}
  \item A relation (after filtering) had too few objects with a desired number of instances (i.e.,~given the relation's statistics, it was impossible to build a balanced answer space within our constraints).
  
 \item We expect the relational information to undergo significant changes (hence, it is not time-invariant). For example, we assume the \textsc{residence} relation, linking an individual to their place of residence, to be highly susceptible to change over time. Including such relations without accounting for their temporal context would unfairly benefit LMs trained on datasets from the same time period as the evaluation data.
  
  \item A relation with many instances that have incomplete object listings in Wikidata since this may cause correct answers to be counted as errors. For example, the \textsc{made-from-material} relation, which connects an object and the material it is made of, often contains only a few primary components as objects. 
  
  \item Overly diverse subject or object space, making the semantics of the relation overly broad and impairing our ability to design meaningful templates. For instance, the relation \textsc{country} connects various entity types, such as events, ships, roles, websites, TLDs, codes of standards, and many more categories, to a country.
\end{itemize}

As a result of this process, we selected 78 relations for inclusion in the larger variant of BEAR.

\subsection{Selecting Relation Instances}

For the selected relations, we retrieve relation instances from Wikidata\footnote{We use the JSON dump of Wikidata of January 3rd, 2022 \citep{wikidataDump2022} which is available as a torrent under a \emph{CC BY-SA} license.} and employ a number of filtering steps to ensure that only instances meeting our desiderata remain.

\paragraph{Filtering subjects and objects.} We first filter down the space of eligible subjects and objects. We remove all Wikidata entities (i.e., subjects and objects) that do not have an English label. Following prior work~\cite{poernerEBERTEfficientYetEffectiveEntity2020}, we additionally remove all subjects with overly revealing entity names. For example, predicting the name of the company that produced the ``Apple Watch'' is straightforward since the correct answer (i.e.,~``Apple'') is part of the subject (i.e.,~``Apple Watch''). The similarity is computed via the token overlap and fuzzy string matching~\cite{bachmannRapidFuzz2023}.

\paragraph{Ensuring a coherent answer space.} Even in our curated set of relations, some relation instances are connected to outlier object types. For instance, the \textsc{head-of-government} relation, which typically connects a country to a specific named person (e.g., ``Joe Biden''), would in some cases connect to a job title instead (e.g., ``president''). 

To increase coherence and ensure that our templates are meaningful, we utilized GPT4 \citep{openaiGPT4TechnicalReport2023} to flag answers which stand out (see Figure~\ref{fig:prompt-answer-space} in Appendix~\ref{sec:used_prompts} for the template that was used)
and decided on a case-by-case basis whether to accept these changes. This process also helped us check the relations for potential issues.

\paragraph{Sampling a balanced dataset.} For this initial set of entities, we sampled the remaining relation instances such that (1) each relation has a uniform distribution of objects in the answer space, (2) each relation has approximately the same number of instances overall, and (3) there is no overlap between various entity names within a relation.
During sampling, we gave preference to Wikidata entities with Wikipedia pages in multiple languages to focus on well-known entities that might reasonably be found in corpora outside of Wikipedia.

This process yields a total of 40,916 instances for our 78 relations.

\subsection{Further Refinement}

The large variant of the dataset consists of a broad range of subjects and captures a considerable amount of relational knowledge found in Wikidata. However, considering the extensive scope of the probe and the evaluation scheme employed, the time required for a complete assessment is considerable. Hence, we further developed a subset of the full probe designed to be more time-efficient.

We imposed a constraint on the answer space, setting an upper limit of 25 answers for 1:N relations. This upper limit was set to keep the task challenging while making the evaluation more efficient. To further refine this subset, we applied a filtering criterion based on entity popularity, excluding entities (objects and subjects) linking to Wikipedia pages with fewer than 10,000 page views (between 2016 and 2023). This step ensures the relevance and recognition of the included entities. Subsequently, through an iterative process, we identified optimal configurations that achieve a balanced answer space with a target of approximately 150 instances. Configurations deemed infeasible were systematically removed (in total, 18 relations, leaving 46 relations). The remaining possible answers were then sorted based on their subjects' median page view count and checked in a manual review. During this review, we examined the answers and their respective subject samples, leading to the exclusion of problematic subject instances. We then included subjects with the highest page view counts. 

In the case of 1-1 relations, the process was similar but using a limit of 60 answers/instances per relation. Additionally, we selected a random sample of instances rather than ranking them by popularity for these relations.

In the following sections of the paper, we will use ``BEAR'' to refer to this subset and ``BEAR\textsubscript{big}'' to denote the full probe.

\subsection{Templates}

We create three templates for each relation to better safeguard against template-specific biases. 
We source the initial templates from the existing LAMA dataset, utilize GPT4 to create additional ones (the used prompt can be found in Figure~\ref{fig:prompt-templates} in Appendix~\ref{sec:used_prompts}), and manually select those that best match our subjects and answer spaces. Finally, we query GPT4 with each of the templates applied to 5 random subject-object pairs from each relation to check for linguistic correctness (the used prompt can be found in Figure~\ref{fig:prompt-instances-with-template} in Appendix~\ref{sec:used_prompts}).

\subsection{Resulting Dataset}

The final dataset consists of 60 relations and 7,731 items. Most of these relations are 1:N, each featuring a restricted answer space ranging from 5 to 25 possible answers, with an average of roughly 23 answers. The answer space is also balanced so that each answer appears the same number of times across all instances. Each answer has between 6 and 30 instances, with an average of approximately 6.5 instances per answer. The dataset also contains 14 1:1 relations that contain only one instance per answer. 

For a detailed comparison of these statistics with LAMA and KAMEL, see Table~\ref{tab:dataset-description}.

\begin{table}
  \small
  \addtolength{\tabcolsep}{-0.15em}
  \begin{tabular}{lrrrr}
    \toprule
    Dataset & LAMA & KAMEL & BEAR\textsubscript{big} & BEAR \\
    \midrule
    \# Instances & 31,479 & 46,800 & 40,916 & 7,731 \\
    \# Relations & 41 & 234 & 78 & 60  \\
    Literals & no & yes & no & no\\
    1:1 Relations & 0 & & 14 & 14 \\
    N:1 Relations & 7 & & 64 & 46 \\
    N:M Relations & 34 & & 0 & 0 \\
    N:M Instances & 1,035 & 4,296 & 0 & 0\\ 
    \makecell{Avg. Instances \\ per Relation} & 830.2 & 1,400\tablefootnote{1,000 train samples, and 200 each for dev \& test} & 596.9 & 149.8\\
    \bottomrule
  \end{tabular}
  \caption{
    Descriptive dataset statistics: BEAR compared to LAMA (T-REx subset) and KAMEL \citep[figures for KAMEL and LAMA from][]{kaloKAMELKnowledgeAnalysis2022}.
    Avg. Instances per Relation only includes relations with more than one instance per answer.
  }\label{tab:dataset-description}
\vspace*{-0.3cm}
\end{table}

\section{Experiments}
\label{sec:experiments}

\begin{table*}[t!]
  \vspace{-0.4cm}
  \centering
  \small
\begin{tabular}{l | crlll}
\toprule
       Model & Type &  \# params & BEAR & BEAR\textsubscript{1:1} & BEAR\textsubscript{1:N}\\
\midrule

Llama-2-13b-hf       & CLM &   13b &  66.9\% \smaller ±  1.0\% &  66.5\% \smaller ±  1.6\% &  67.0\% \smaller ±  1.1\% \\
Mistral-7B-v0.1      & CLM &  7.0b &  65.4\% \smaller ±  1.1\% &  64.5\% \smaller ±  1.2\% &  65.5\% \smaller ±  1.1\% \\
gemma-7b             & CLM &  7.0b &  63.7\% \smaller ±  1.3\% &  63.5\% \smaller ±  0.7\% &  63.8\% \smaller ±  1.4\% \\
Llama-2-7b-hf        & CLM &  7.0b &  62.4\% \smaller ±  1.3\% &  62.2\% \smaller ±  1.1\% &  62.4\% \smaller ±  1.3\% \\
gemma-2b             & CLM &  2.0b &  51.5\% \smaller ±  1.0\% &  53.1\% \smaller ±  1.3\% &  51.3\% \smaller ±  1.0\% \\
opt-30b              & CLM &   30b &  47.9\% \smaller ±  0.5\% &  45.8\% \smaller ±  1.0\% &  48.2\% \smaller ±  0.6\% \\
opt-13b              & CLM &   13b &  45.4\% \smaller ±  0.8\% &  43.5\% \smaller ±  2.1\% &  45.7\% \smaller ±  0.6\% \\
opt-6.7b             & CLM &  6.7b &  43.8\% \smaller ±  1.1\% &  42.5\% \smaller ±  1.0\% &  43.9\% \smaller ±  1.2\% \\
opt-2.7b             & CLM &  2.7b &  37.3\% \smaller ±  0.9\% &  35.6\% \smaller ±  0.7\% &  37.5\% \smaller ±  1.0\% \\
opt-1.3b             & CLM &  1.3b &  31.5\% \smaller ±  0.8\% &  31.3\% \smaller ±  0.6\% &  31.5\% \smaller ±  0.9\% \\
gpt2-xl              & CLM &  1.6b &  26.2\% \smaller ±  0.7\% &  24.1\% \smaller ±  1.6\% &  26.5\% \smaller ±  0.6\% \\
gpt2-large           & CLM &  812M &  22.2\% \smaller ±  0.6\% &  20.1\% \smaller ±  1.8\% &  22.5\% \smaller ±  0.5\% \\
roberta-large        & MLM &  355M &  21.5\% \smaller ±  0.8\% &  22.0\% \smaller ±  1.1\% &  21.5\% \smaller ±  0.8\% \\
bert-large-cased     & MLM &  335M &  19.9\% \smaller ±  0.5\% &  16.6\% \smaller ±  1.0\% &  20.3\% \smaller ±  0.5\% \\
opt-350m             & CLM &  350M &  19.6\% \smaller ±  0.6\% &  18.6\% \smaller ±  1.2\% &  19.7\% \smaller ±  0.6\% \\
gpt2-medium          & CLM &  355M &  19.0\% \smaller ±  0.8\% &  16.0\% \smaller ±  2.6\% &  19.4\% \smaller ±  0.6\% \\
bert-base-cased      & MLM &  109M &  18.4\% \smaller ±  0.4\% &  15.0\% \smaller ±  1.1\% &  18.8\% \smaller ±  0.4\% \\
roberta-base         & MLM &  125M &  16.4\% \smaller ±  0.7\% &  15.8\% \smaller ±  1.8\% &  16.5\% \smaller ±  0.8\% \\
opt-125m             & CLM &  125M &  16.4\% \smaller ±  0.5\% &  14.0\% \smaller ±  1.3\% &  16.7\% \smaller ±  0.4\% \\
xlm-roberta-large    & MLM &  561M &  14.3\% \smaller ±  0.3\% &  14.9\% \smaller ±  1.7\% &  14.3\% \smaller ±  0.5\% \\
gpt2                 & CLM &  137M &  13.5\% \smaller ±  0.8\% &   9.4\% \smaller ±  2.1\% &  14.0\% \smaller ±  0.7\% \\
xlm-roberta-base     & MLM &  279M &  11.4\% \smaller ±  0.2\% &  11.4\% \smaller ±  1.1\% &  11.4\% \smaller ±  0.2\% \\
Random Baseline      &   - &     - &   4.7\%                   &   1.7\%                   &   5.1\%                  \\
\bottomrule
  \end{tabular}
  \caption{Models investigated in this work \cite{devlinBERTPretrainingDeep2019b, jiangMistral7B2023, geminiteamGeminiFamilyHighly2023, liuRoBERTaRobustlyOptimized2019, radfordLanguageModelsAre2019, touvronLlamaOpenFoundation2023, zhangOPTOpenPretrained2022} sorted by their BEAR score (weighted average over all relations) and as the mean over all templates (with the standard error).}
  \label{tab:model-statistics}
  \vspace{-0.3cm}
\end{table*}

We present an experimental evaluation using the BEAR probe on a selection of LMs, compare our results to earlier probes, and discuss the results.

\paragraph{Compared LMs.} We compare a total of 22 LMs, as listed in Table~\ref{tab:model-statistics}: This includes 6 masked LMs from the BERT, RoBERTa, XLM-RoBERTa families, each in their \texttt{base} and \texttt{large} variants. Additionally, we include 16 causal LMs from the GPT and OPT families, along with newer models such as Llama2, Gemini, and Mistral. We assess 5 different sizes for the OPT models to examine the relationship between the BEAR score and increasing model sizes.

\paragraph{BEAR score.} We compute the BEAR score for each of the three template options per relation individually and report the average across templates as well as the standard deviation.

\subsection{Main Results}

\begin{figure}[t!]
    \centering
    \includegraphics[width=\linewidth, clip, trim={0cm 0cm 0cm 0.7cm}]{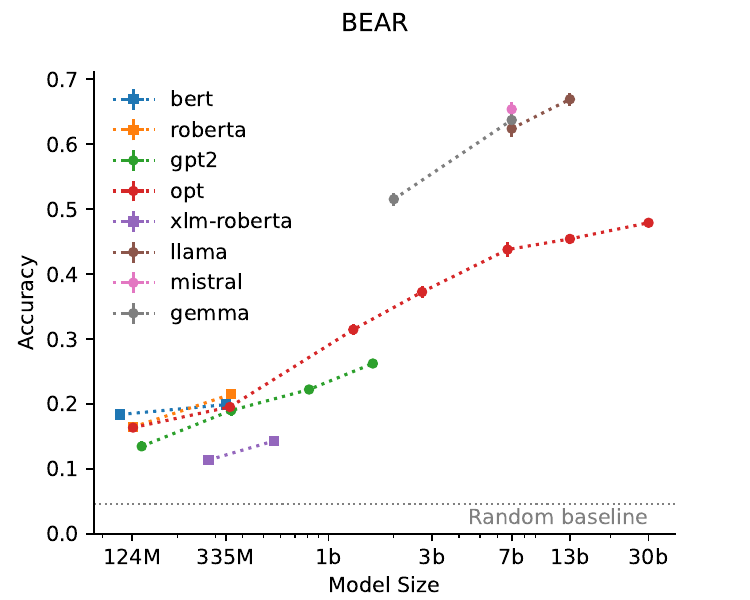}
    \caption{Probing scores of different models on BEAR. Model size is represented on a log scale.}
    \label{fig:accuracy_by_model_size}
\vspace*{-0.2cm}
\end{figure}

Table~\ref{tab:model-statistics} lists the results for all LMs in consideration. We present the overall BEAR score and the scores for the subsets of 1:1 and N:1 relations only.  We find that scores are generally low for all models, highlighting the challenging nature of our benchmark, as it queries factual information with strong detractors. 
In addition, we make a number of observations:

\paragraph{BEAR scores are higher for larger LMs.} In line with our expectations, we find that larger models consistently outperform their smaller counterparts. For a better illustration, we present a plot of accuracy against model size in Figure~\ref{fig:accuracy_by_model_size}. This trend of steady accuracy improvement with increasing model size is evident across all tested model families. Interestingly, the smallest change is observed among BERT models, where the performance of \texttt{bert-base-cased} and \texttt{bert-large-cased} across all of the relations is roughly on par. We further observe that recent models generally achieve higher performance, with \texttt{Mistral-7B-v0.1} (2023) and \texttt{gemma-7b} (2024) surpassing the \texttt{Llama2-7b} (2023) model (when compared at identical parameter counts), which itself significantly outperforms the \texttt{opt-6.7b} model from 2022.

\noindent\textbf{Better BEAR scores for masked LMs.} When comparing models by their parameter count, we note a slight advantage of masked over causal LMs. This may indicate that the masked language modeling objective, encouraging deep bidirectionality, is more effective in capturing factual knowledge. 

\noindent\textbf{Impact of multilingual training data.} We note that the two XLM-RoBERTa models are among the lowest-scoring models in the benchmark. We hypothesize that this diminished performance of the XLM models may stem from its pre-training on multilingual corpora and a focus of BEAR on English-language entities. 

\noindent\textbf{Impact of templates.} We further evaluate the impact of template choice on the BEAR score. A full analysis of all relations is provided in Figure~\ref{sec:more_results} in the Appendix. 

In line with the observation of \citet{elazarMeasuringImprovingConsistency2021}, we find that LMs are sensitive to how they are queried. For instance, in the \textsc{capital-of} relation, the accuracy of \texttt{bert-base-cased} drops approximately by 80\% when using ``[Y] has its governmental seat in [X]'' instead of ``The capital of [X] is [Y].''. While some states have their government seat in a different location than their official capital city, this fact alone cannot account for the significant disparity in observed performance. Interestingly, we don't observe any drop in performance between these templates in the case of other models (e.g., Llama2 models). We hypothesize that this might be due to BERT's primary training on Wikipedia, which has limited exposure to diverse writing styles.

\subsection{Comparison to LAMA}

\begin{figure}
    \centering
    \includegraphics[width=\linewidth]{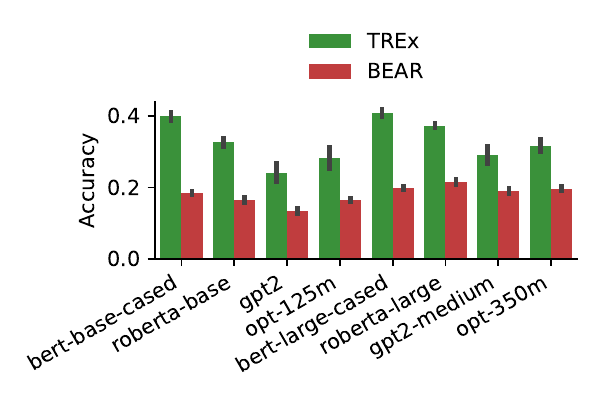}
    \caption{Comparative analysis of model performance on identical subsets of relations and templates in the T-REx (LAMA) and BEAR datasets using the log-likelihood based evaluation.}
    \label{fig:bear_vs_trex_subset}
\end{figure}

To compare BEAR and LAMA, we consider their common subset of relations and utilize our proposed log-likelihood based evaluation technique. See Figure~\ref{fig:bear_vs_trex_subset} for an illustration. We find BEAR to be a more challenging probe compared to T-REx, the dataset used in LAMA. We attribute this to several design choices, namely the balanced answer space and the absence of overly informative entity names, forcing LMs to rely solely on the knowledge encoded within its parameters.  

We also find that the performance disparity evaluated on BEAR's relations is less pronounced than across the same subset of T-REx's relations. Such difference may indicate that models pre-trained on Wikipedia (like BERT) have an advantage on LAMA over those not trained on Wikipedia (such as GPT2) due to a potential train/test data overlap. For a detailed performance comparison on a per-relation basis, refer to Figure~\ref{fig:bear_vs_trex_bbc_all_relations} in the Appendix. For example, \texttt{bert-base-cased} achieves a very high performance on T-REx's \textsc{manufacturer} relation (ID: P176) but a significantly lower score on the corresponding BEAR subset.

\subsubsection{Ablations}

\begin{figure}[t!]
    \centering
    \includegraphics[width=\linewidth]{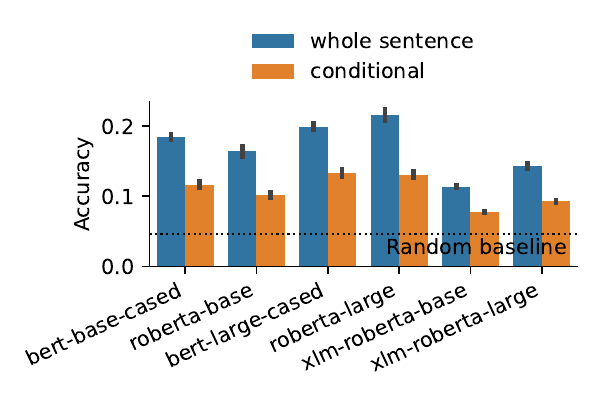}
    \caption{Aggregated accuracy (measured on BEAR) when using the sum over all tokens in the complete statement vs. answer-tokens only.}
    \label{fig:mlm-conditional_vs_whole}
\end{figure}

\noindent\textbf{Conditional scores.}
To compute the pseudo log-likelihood of a statement in masked language models, one forward pass per token is required. However, masking only the tokens that are part of the answer would significantly reduce the required computation. One may expect that the representation of the answer tokens may be sufficient to predict the likelihood of each answer.
Still, our experiments (see Figure~\ref{fig:mlm-conditional_vs_whole}) indicate there is a significant\footnote{
P-value of \num{1.1e-11}; using a Student's t-test for paired samples
} drop in performance when using the conditional score (i.e., the score of the answer tokens conditioned on the context instead of the pseudo log-likelihood of the complete sentence). Coincidentally, for entities represented by a single token, the conditional score matches the score predicted for the [MASK] token, similar to the approach first used in the LAMA probe.

\noindent\textbf{Sum vs.~mean of the log-likelihood.}
\label{sec:sum-mean-section} We investigate how performance varies by scoring the sentences using both sum and mean reduction methods.
We observe that scoring the sentence by computing the mean over the token log-likelihoods tends to yield inferior performance for the probe. Figure~\ref{fig:sum_vs_mean} (in Appendix~\ref{sec:more_results}) illustrates the results of this ablation study. 

To understand why this might be the case, consider how the word 'souvenir' can be broken down into subword tokens: `so', `\#\#uven', and `\#\#ir'.\footnote{
The example was introduced by \citet{kaufBetterWayMasked2023b} to make a different point but is also relevant here.}
The first token `so' may have a relatively small log-likelihood, `\#\#uven' a bit higher (since it's conditioned on the first token), and finally `\#\#ir' a log-likelihood of almost 0 since there are few other ways to continue the statement. In contrast, the single-token word `gift' may have a medium log-likelihood, which may still be lower than the average of the three-token word `souvenir'. This example illustrates that the mean can inflate the predicted probabilities of longer answers or sentences.

We suggest summating the tokens' log-likelihoods for both masked and casual language models in future experiments.

\section{Conclusion}

We presented BEAR, a relational knowledge probe applicable to both causal and masked LMs. Since our proposed approach imposes no restrictions on the evaluation data, we created a large evaluation dataset that addresses issues of answer skews, domain and template bias, and the correctness of facts identified by ourselves and prior work. We publicly release BEAR for use by the research community.  

\section*{Limitations and Risks}

The knowledge probe we present in this paper follows the approach of earlier probes and, as such, tests only for factual, relational knowledge. This kind of knowledge includes classic relationship types such as the place of birth of persons, their time of birth, the genre of works of art, etc. However, one might be interested in testing a model for other types of more general commonsense knowledge, such as physical reasoning or general properties of concepts. Our probe does not test for such kinds of knowledge. 

Furthermore, even though we devised heuristics to ensure that entities in BEAR are common enough to appear on Wikipedia pages of many different languages and record a certain number of page views, there remains a likely bias towards entities overrepresented on Wikipedia, giving an advantage to LMs trained on Wikipedia rather than more general corpora. 

We see few risks in the BEAR probe itself but caution that knowledge probing is often used to assist in the research and development of LMs. As such, BEAR may contribute to developing LMs that malevolent actors might misuse.

\section*{Acknowledgements}

We thank all reviewers for their valuable comments.
Jacek Wiland, Max Ploner, and Alan Akbik are supported by the Deutsche Forschungsgemeinschaft (DFG, German Research Foundation) under Germany’s Excellence Strategy – EXC 2002/1 “Science of Intelligence” – project number 390523135. Alan Akbik is further supported by the Deutsche Forschungsgemeinschaft (DFG, German Research Foundation) under the Emmy Noether grant ``Eidetic Representations of Natural Language'' (project number 448414230)

\bibliography{anthology,custom}

\appendix

\clearpage

\section{Further Results}\label{sec:more_results}

\begin{figure}[h]
    \centering
    \includegraphics[width=\linewidth, trim={0cm 0.2cm 0cm 0cm},]{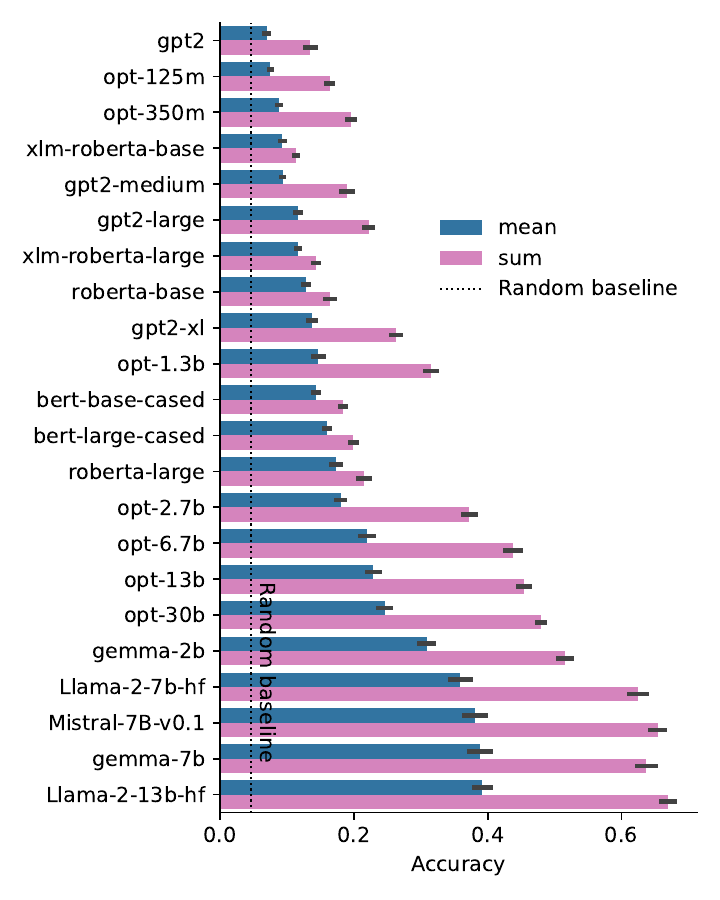}     
    \caption{Aggregated accuracy of different retrieval variants on BEAR. The error bars indicate the standard error over three evaluations using the different templates.}
    \vspace{-0.3cm}
    \label{fig:sum_vs_mean}
\end{figure}

\subsection{Results on BEAR\textsubscript{big}}

In addition to the accuracy on the further refined BEAR dataset, we report the scores on BEAR\textsubscript{big} (see Table~\ref{tab:results-bear-big}). The ranking is very similar and differs only in a few positions.

\begin{table*}
  \vspace{-0.4cm}
  \centering
  \small
\begin{tabular}{l | crlll}
\toprule
       Model & Type &  \# params & BEAR & BEAR\textsubscript{1:1} & BEAR\textsubscript{1:N}\\
\midrule

Llama-2-13b-hf       & CLM &   13b &  42.0\% \smaller ±  0.6\% &  54.3\% \smaller ±  1.3\% &  41.2\% \smaller ±  0.6\% \\
Mistral-7B-v0.1      & CLM &  7.0b &  41.0\% \smaller ±  0.6\% &  52.6\% \smaller ±  1.2\% &  40.2\% \smaller ±  0.6\% \\
gemma-7b             & CLM &  7.0b &  39.5\% \smaller ±  0.8\% &  52.0\% \smaller ±  1.1\% &  38.6\% \smaller ±  0.8\% \\
Llama-2-7b-hf        & CLM &  7.0b &  37.5\% \smaller ±  0.8\% &  50.3\% \smaller ±  1.1\% &  36.6\% \smaller ±  0.7\% \\
gemma-2b             & CLM &  2.0b &  29.1\% \smaller ±  0.6\% &  41.4\% \smaller ±  1.1\% &  28.2\% \smaller ±  0.6\% \\
opt-30b              & CLM &   30b &  25.6\% \smaller ±  0.3\% &  35.7\% \smaller ±  0.9\% &  24.9\% \smaller ±  0.3\% \\
opt-13b              & CLM &   13b &  24.2\% \smaller ±  0.5\% &  33.6\% \smaller ±  1.6\% &  23.5\% \smaller ±  0.5\% \\
opt-6.7b             & CLM &  6.7b &  23.2\% \smaller ±  0.7\% &  33.1\% \smaller ±  0.5\% &  22.5\% \smaller ±  0.7\% \\
opt-2.7b             & CLM &  2.7b &  19.3\% \smaller ±  0.4\% &  27.6\% \smaller ±  0.8\% &  18.7\% \smaller ±  0.3\% \\
opt-1.3b             & CLM &  1.3b &  16.0\% \smaller ±  0.4\% &  23.3\% \smaller ±  1.0\% &  15.5\% \smaller ±  0.4\% \\
gpt2-xl              & CLM &  1.6b &  12.9\% \smaller ±  0.2\% &  17.8\% \smaller ±  1.2\% &  12.6\% \smaller ±  0.2\% \\
roberta-large        & MLM &  355M &  11.1\% \smaller ±  0.4\% &  17.1\% \smaller ±  0.8\% &  10.7\% \smaller ±  0.4\% \\
gpt2-large           & CLM &  812M &  10.7\% \smaller ±  0.2\% &  14.0\% \smaller ±  1.5\% &  10.5\% \smaller ±  0.2\% \\
bert-large-cased     & MLM &  335M &  10.1\% \smaller ±  0.3\% &  11.8\% \smaller ±  0.7\% &  10.0\% \smaller ±  0.3\% \\
bert-base-cased      & MLM &  109M &   9.6\% \smaller ±  0.3\% &  11.5\% \smaller ±  1.2\% &   9.4\% \smaller ±  0.3\% \\
opt-350m             & CLM &  350M &   9.5\% \smaller ±  0.2\% &  13.4\% \smaller ±  0.8\% &   9.2\% \smaller ±  0.2\% \\
gpt2-medium          & CLM &  355M &   9.1\% \smaller ±  0.3\% &  11.3\% \smaller ±  1.9\% &   8.9\% \smaller ±  0.2\% \\
roberta-base         & MLM &  125M &   8.4\% \smaller ±  0.3\% &  11.8\% \smaller ±  1.8\% &   8.1\% \smaller ±  0.4\% \\
opt-125m             & CLM &  125M &   8.0\% \smaller ±  0.2\% &   9.5\% \smaller ±  0.8\% &   7.9\% \smaller ±  0.2\% \\
xlm-roberta-large    & MLM &  561M &   7.4\% \smaller ±  0.2\% &  11.2\% \smaller ±  1.6\% &   7.1\% \smaller ±  0.1\% \\
gpt2                 & CLM &  137M &   6.4\% \smaller ±  0.3\% &   5.8\% \smaller ±  1.6\% &   6.5\% \smaller ±  0.2\% \\
xlm-roberta-base     & MLM &  279M &   5.8\% \smaller ±  0.1\% &   8.7\% \smaller ±  0.9\% &   5.6\% \smaller ±  0.0\% \\
Random Baseline      &   - &     - &   2.5\%                   &   0.5\%                   &   2.7\%                  \\

\bottomrule
  \end{tabular}
  \caption{Models investigated in this work evaluated on BEAR\textsubscript{big} (weighted average over all relations) and as the mean over all templates (with the standard error).}
  \label{tab:results-bear-big}
  \vspace{-0.3cm}
\end{table*}

\subsection{Ablation: Pseudo log-likelihood metric}
\begin{figure}[h]
    \centering
    \includegraphics[width=\linewidth]{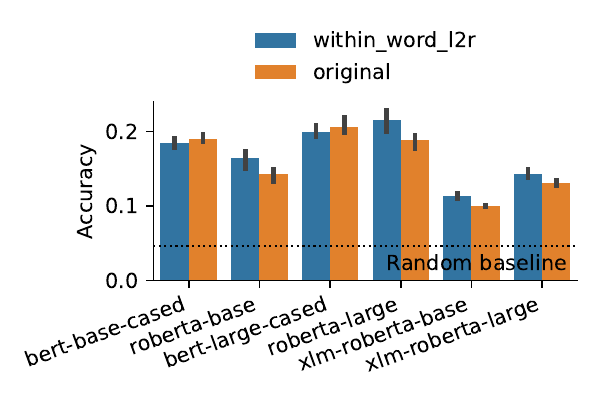}     
    \caption{Aggregated accuracy of different retrieval variants on BEAR. The error bars indicate the standard error over three evaluations using the different templates.}
    \label{fig:mlm-pll-metrics}
\end{figure}

While in preliminary experiments on LAMA, we observed a higher benefit from using the 
\texttt{within\_word\_l2r} variant, it has only a slightly higher mean score than \texttt{original} (see Figure~\ref{fig:mlm-pll-metrics}).
This difference is not significant when using the \texttt{sum} variant (p-value of \num{0.52} on a Student's t-test for paired samples).
However, the difference is large when using the \texttt{mean} variant (and significant with p-value of \num{0.025})

\subsection{Impact of Pre-Training Data on the BEAR Score}
\label{sec:new-knowledge} In order to verify the actual ability of the BEAR probe to measure knowledge contained in models' pre-trained weights, we set up an ablation study. We hypothesize that models trained solely on domain-specific datasets, without exposure to the general knowledge tested by BEAR, will show significantly reduced performance compared to those trained on more general datasets, given that the models have similar architectures and sizes.
A family of BioGPT models \citep{10.1093/bib/bbac409} was based on the GPT-2 architecture with the sole differences arising from the pre-training data: specifically, they were trained on PubMed abstracts and titles rather than on data from web crawls as it was the case for the GPT models. The evaluation results confirm our hypothesis, with the BioGPT model achieving an average score of 10.84\% and its large variant reaching 13.6\% on BEAR, both trailing behind the \texttt{gpt2-medium} (19.0\%) and \texttt{gpt2-xl} (26.2\%) models.

\begin{figure}
    \centering
    \includegraphics[width=\linewidth]{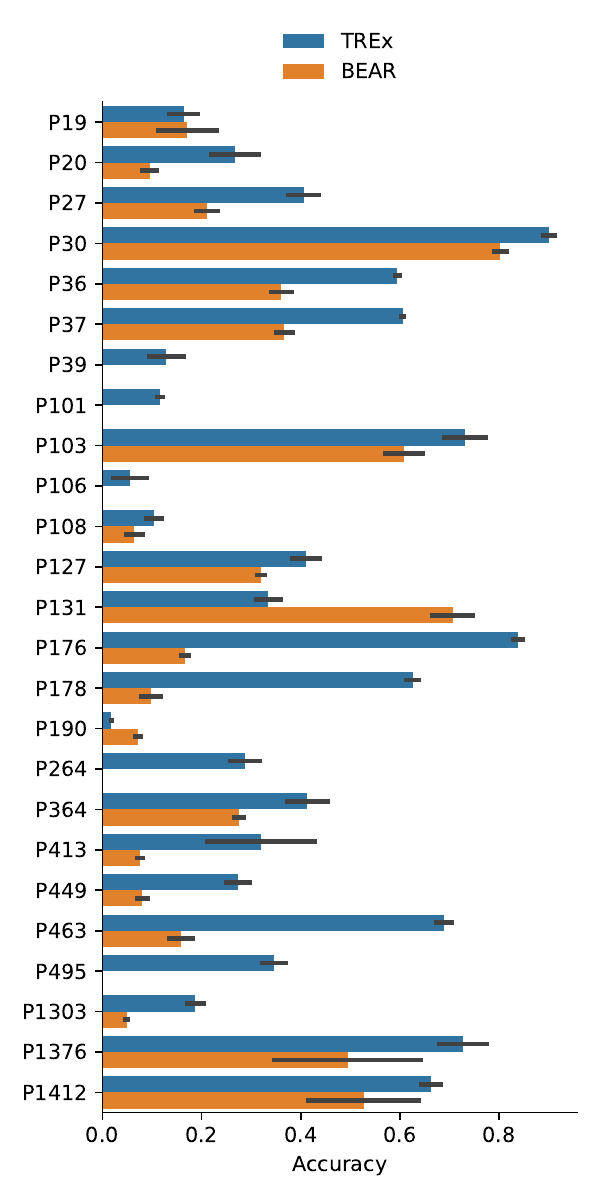}
    \caption{Performance of \texttt{bert-base-cased} on a per-relation basis for both BEAR and T-REx probes. The results were obtained by summing pseudo log-likelihood scores (\texttt{within\_word\_l2r})}
    \label{fig:bear_vs_trex_bbc_all_relations}
\end{figure}

\begin{figure}
    \centering
    \begin{subfigure}{\linewidth}
        \includegraphics[width=\linewidth, trim={0cm 0cm 0cm 0.7cm}, clip]{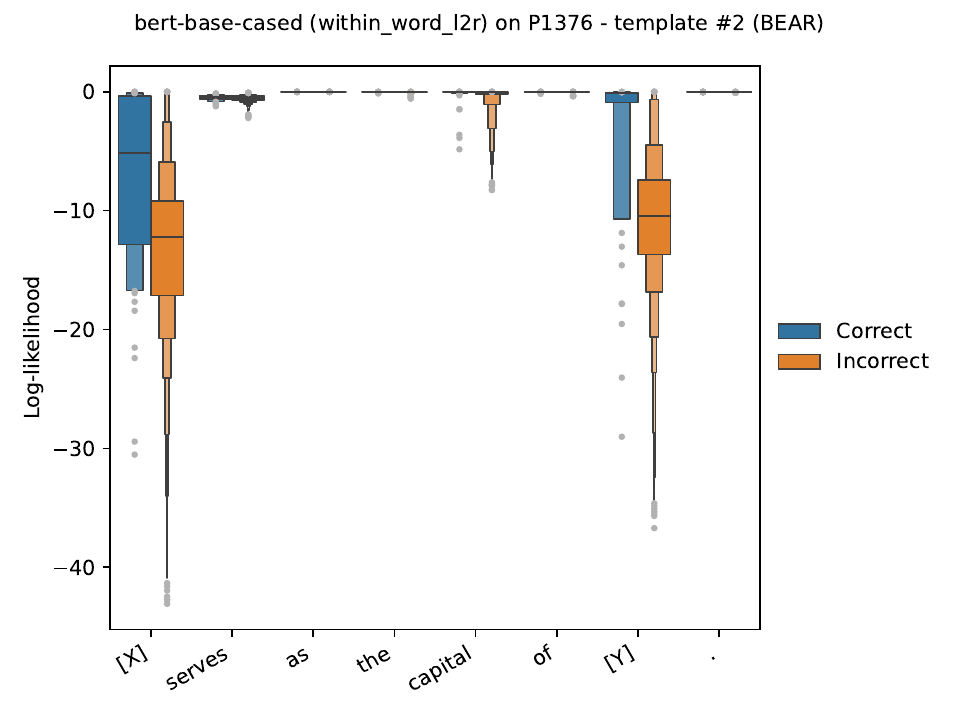}
        \caption{First Template: ``[X] is the capital of [Y].''; Accuracy of 63\%}
    \end{subfigure}
    \begin{subfigure}{\linewidth}
        \includegraphics[width=\linewidth, trim={0cm 0cm 0cm 0.7cm}, clip]{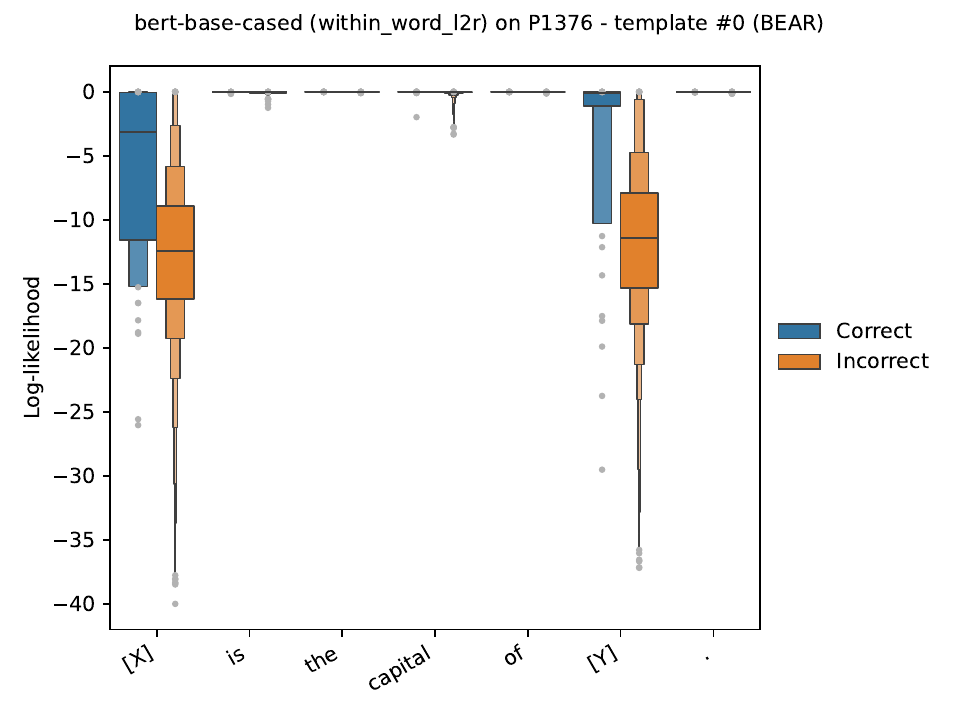}
        \caption{Second Template: ``[Y] has its governmental seat in [X].''; Accuracy of 20\%}
    \end{subfigure}
        \begin{subfigure}{\linewidth}
        \includegraphics[width=\linewidth, trim={0cm 0cm 0cm 0.7cm}, clip]{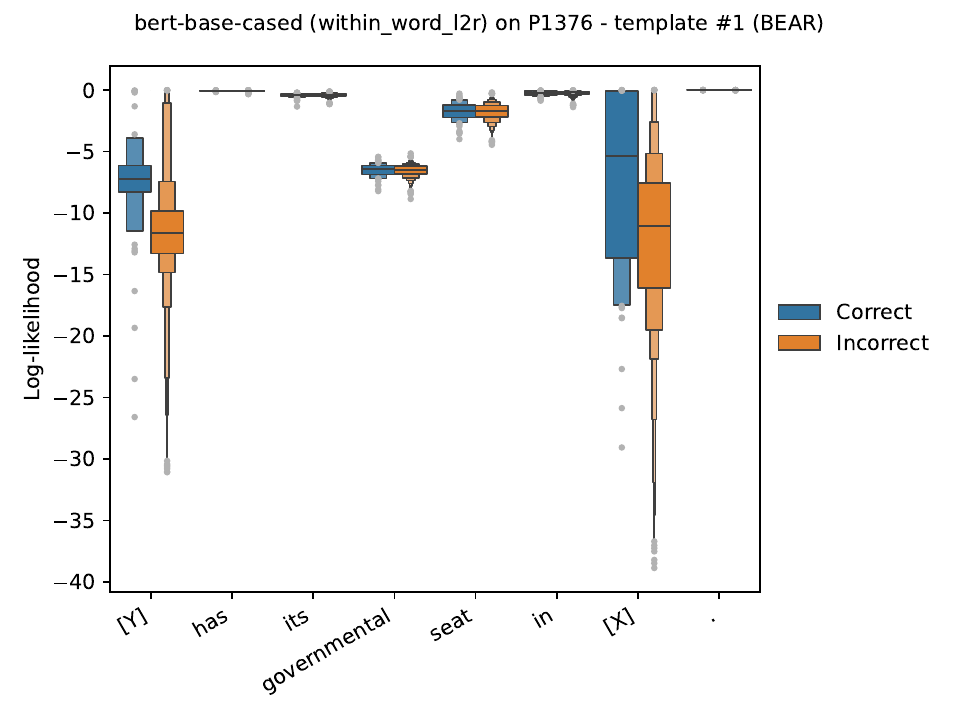}
        \caption{Third Template: ``[X] serves as the capital of [Y].''; Accuracy of 65\%}
    \end{subfigure}
    \caption{\texttt{bert-base-cased} on $\mathrm{P1376}$ (BEAR)}
    \label{fig:token_scores}
\end{figure}

\section{Noise Levels}\label{sec:noise-levels}

Crowdsourced knowledge bases like Wikidata often contain noisy and inaccurate data. To reduce this issue, we developed heuristics to select better-known entities that are more likely to be verified and corrected. Nonetheless, the potential for noise leakage exists. To evaluate the extent of such noise, we performed a noise levels analysis, in which we manually reviewed 100 randomly chosen examples from various relations for both BEAR\textsubscript{big} and LAMA (T-REx) probes. We cross-validated the accuracy of the information from Wikidata with alternative sources, confirming 96\% and 97\% of examples for BEAR\textsubscript{big} and T-REx, respectively\footnote{A two-sided z-test yields a p-value of 0.6698}. 

Additionally, we evaluated whether the object of the relation truly represents the only correct answer as our benchmark (expecting a single correct answer) would mark those answers as incorrect. About 11\% of BEAR's answers included multiple correct responses within its answer range, significantly lower than T-REx's 35\%\footnote{A two-sided z-test yields a p-value of 0.0001}, which uses BERT's vocabulary as its answer space. 

Thus, while BEAR\textsubscript{big} and LAMA exhibit similar noise levels, BEAR\textsubscript{big} demonstrates higher reliability by effectively reducing the incidence of multiple correct answers.
Moreover, the refined subset of BEAR, due to additional filtering steps, is expected to decrease these values even further.

\section{GPU Compute Time}\label{sec:gpu-compute}

Due to the diversity of GPU hardware utilized in our experiments, we have chosen not to present cumulative GPU hours. Instead, to provide insights into the computational requirements of our research, we offer a selection of representative examples highlighting individual experiments. The duration for the BEAR probe evaluations notably differed across models. For instance, evaluating the \texttt{opt-6.7b} model on the larger variant of the probe using the NVIDIA A100 (80GB) machine required approximately 14 hours, whereas \texttt{gpt2} averaged about 1 hour. Additionally, causal language models were evaluated faster due to their method for calculating sentence log-likelihoods. For instance, evaluating the \texttt{bert-base-cased} model, roughly the same size as \texttt{gpt2}, on the entire BEAR probe took four times as long. 

Moreover, as stated in the main text, assessing the large variant of BEAR requires a significant amount of time. For example, \texttt{bert-base-cased} requires almost 4.5 hours when evaluating the BEAR\textsubscript{big} probe with NVIDIA RTX 3090. Yet, this duration drops to around 16 minutes when using the smaller BEAR probe on the same hardware.

\section{Predictions' Confidence}\label{sec:predictions-confidence}
Log-likelihoods assigned to textual statements by language models over a closed set of answers can be converted into values that can be interpreted as \textit{probabilities}. Here, the idea involves transforming these negative values with the softmax function to obtain a normalized set of values that together resemble a probability distribution. These scores reflect the model's confidence in each possible answer within the set. 
Furthermore, by comparing the calculated probability distribution with a uniform distribution, we can derive a \textit{uncertainty score}, which reflects the model's confusion for a given instance. This score is based on the Kullback-Leibler Divergence between the predicted probabilities and the uniform distribution. Specifically, we derive divergence between a predicted probability distribution $P$ and a uniform distribution $U$ over N possible answers (outcomes). We then normalize this score to make it scale-free and bounded between 0 and 1 (entropy is maximal for the uniform distribution). Finally, we take a complement of the obtained value (by subtracting it from 1) effectively reversing its interpretation, shifting the focus from divergence to similarity to the uniform distribution. The mathematical formulation of this metric can be found in Equations~\ref{eq:KL_divergence}, \ref{eq:normalized_KL}, \ref{eq:uncertainty}.

\begin{equation}
\begin{aligned}
    KL(P \|U) =
    =& \sum_{i=0}^N p_i \log \frac{p_i}{\nicefrac{1}{N}} \\
    =& \sum_{i=0}^N p_i \log p_i + \sum_{i=0}^N p_i \log N \\
    =& \log N - H(P) \\
    =& H(U) - H(P)  \\
    & \in [0, H(U)]
\end{aligned}
\label{eq:KL_divergence}
\end{equation}

\begin{equation}
\begin{aligned}
&\text{Normalized KL} = 1 - \frac{H(P)}{H(U)} \in [0,1]
\end{aligned}
\label{eq:normalized_KL}
\end{equation}

\begin{equation}
\begin{aligned}
\text{Uncertainty} &= 1 - \text{Normalized KL} \\
&= \frac{H(P)}{H(U)} \in [0,1].
\end{aligned}
\label{eq:uncertainty}
\end{equation}

The interpretation of such a score is as follows: we observe the largest uncertainty when the predicted distribution approximates a uniform distribution, and the lowest when the entropy of the predicted distribution is null. The latter situation occurs with a point mass distribution, which allocates all the probability mass to one outcome.

\section{Error Analysis}

We further conducted an error analysis on the mispredictions generated by the  \texttt{Llama-2-13b-hf} model. The investigation did not reveal the presence of any systematic categories or clusters of errors. Some errors appeared to be similar to educated guesses. For instance, \texttt{Llama-2-13b-hf} outputs the highest log-likelihood for the factually incorrect statement such as ``Kazım Ayvaz passed away in Turkey''. However, this is not an unreasonable assumption, given that \textit{Kazım Ayvaz} was a Turkish Olympic medalist born in Turkey. Even though the correct answer is \textit{Sweden}, this illustrates that some of the model's errors may stem from logical assumptions instead of arbitrary mistakes. 

On the other hand, certain errors appear to be random, with the predicted answer having no apparent connection to the subject. For example, \texttt{Llama-2-13b-hf} assigned the highest score to the incorrect answer of \textit{Massif Central} (highlands in France) when queried about the location of the Welsh mountain \textit{Elidir Fawr}. 

Furthermore, on rare occasions, the accuracy of the model's predictions is compromised by issues such as noise leakage or the imperfect quality of data sourced from Wikidata (for the analysis of such noise, see Section ~\ref{sec:noise-levels} in the Appendix). For instance, a Wikidata entry mentioning \textit{Ronaldo} refers to Ronaldo Luís Nazário de Lima, commonly known simply as Ronaldo, a well-known player for the Brazilian national football team. However, the model erroneously identifies \textit{Ronaldo} as Cristiano Ronaldo, a renowned Portuguese footballer.  Table ~\ref{tab:error-analysis} presents a selection of errors made by the  \texttt{Llama-2-13b-hf} model and provides further information on prediction's confidence as described in Section \ref{sec:predictions-confidence}. Specifically, it gives the ranks of the correct and predicted statements as assessed by their log-likelihoods as well as their probability scores obtained by applying the softmax function over all log-likelihoods in the rankings.

\section{Prompts Used}\label{sec:used_prompts}

All prompts were passed as `system messages` to Chat-GPT4 API and are presented in Figures ~\ref{fig:prompt-answer-space}, ~\ref{fig:prompt-templates}, and ~\ref{fig:prompt-instances-with-template}.

\begin{figure*}
    \centering
    \setlength\fboxsep{1em}
    \fbox{%
    \parbox{0.9\linewidth}{%
        \ttfamily\smaller
            You are a researcher assistant tasked to design an evaluation dataset to test relational knowledge contained in language models. Specifically, you are given a label for a relation, its description, and a list of possible answers. Your assignment is to identify words that do not align with the majority category in a given list of answers given the relation label and its description. Return your response as a Python tuple. The first element should be a list containing the words that don't fit the majority category, and the second element should be a string representing the category of the majority of answers. If all words fit the category, return an empty list. Example format: (['Berlin', 'Warsaw'], 'countries').
        }%
    }
    \caption{Prompt used to flag words in the answer space of each relation. In addition to some relation metadata (label and description) the (intermediate) answer space was passed on the model.}
    \label{fig:prompt-answer-space}
\end{figure*}

\begin{figure*}
    \centering
    \setlength\fboxsep{1em}
    \fbox{%
    \parbox{0.9\linewidth}{%
            \ttfamily\smaller
            As a research assistant, your task is to create an evaluation dataset to assess the relational knowledge of language models. You are provided with a specific relation label, its definition, and examples of subjects and objects related to it. Your objective is to craft three semantically similar cloze sentence templates that embody this relation. Use '[X]' as a placeholder for the subject and '[Y]' for the object (answer). Ensure that these sentence templates are straightforward and devoid of superfluous elements. For instance, given {'label': 'educated at', 'description': 'educational institution attended by subject', 'subjects': ['Einstein', 'Feynman'], 'objects': ['Princeton University', 'University of Zurich']}, your templates might be: ['[X] was educated at [Y].', '[X] studied at [Y].', '[X] was a student at [Y].']. Present your response as a Python list.
        }%
    }
    \caption{Prompt used to generate new template variants. Alongside the relation metadata, including label and description, 6 subject-object pairs were provided as examples for each relation.}
    \label{fig:prompt-templates}
\end{figure*}

\begin{figure*}
    \centering
    \setlength\fboxsep{1em}
    \fbox{%
    \parbox{0.9\linewidth}{%
        \ttfamily\smaller
            Evaluate the linguistic correctness of the following sentence. If it is correct, return 'Correct'. If incorrect, identify the error and suggest a revised sentence. For instance, 'I used to live in USA.' ->  'I used to live in the USA.', 'My name is John.' ->  'Correct'.
        }%
    }
    \caption{System message for GPT-4 API call used to identify potential problems with the templates for each relation. We assessed all three templates by populating them with 5 random subject-object pairs. Although the prompt was intended to detect linguistic issues in the templates, it also facilitated the discovery of further issues with the relation instances.}
    \label{fig:prompt-instances-with-template}
\end{figure*}

\begin{table*}[t!]
  \vspace{-0.4cm}
\centering
\addtolength{\tabcolsep}{-0.2em}
{\scriptsize 
\begin{tabular}{lcclccc}
\toprule
\small Relation & \small Index &  \small Correct & \small Instance & \small Rank & \small Probability & \small Uncertainty \\
\midrule
    \multirow{2}{*}{P20} & \multirow{2}{*}{65} & False & \textcolor{subblue}{Kazım Ayvaz} passed away in \textcolor{objorange}{Turkey}. & 1 & 0.6038 & \multirow{2}{*}{0.3963} \\
 &  & True & \textcolor{subblue}{Kazım Ayvaz} passed away in \textcolor{objorange}{Sweden}. & 5 & 0.0132 &  \\
\midrule
    \multirow{2}{*}{P69} & \multirow{2}{*}{65} & False & \textcolor{subblue}{Gary Trent Jr.} was educated at the \textcolor{objorange}{University of Washington}. & 1 & 0.7873 & \multirow{2}{*}{0.2573} \\
 &  & True & \textcolor{subblue}{Gary Trent Jr.} was educated at the \textcolor{objorange}{Duke University}. & 2 & 0.0814 &  \\
\addlinespace[2pt] \midrule \addlinespace[2pt]
\multirow{2}{*}{P509} & \multirow{2}{*}{284} & False & \textcolor{subblue}{Camille Pissarro} died from \textcolor{objorange}{tuberculosis}. & 1 & 0.4592 & \multirow{2}{*}{0.5945} \\
 &  & True & \textcolor{subblue}{Camille Pissarro} died from \textcolor{objorange}{sepsis}. & 2 & 0.1854 &  \\
\addlinespace[2pt] \midrule \addlinespace[2pt]
\multirow{2}{*}{P19} & \multirow{2}{*}{6} & False & \textcolor{subblue}{Didier Marouani} was born in \textcolor{objorange}{Belgium}. & 1 & 0.4284 & \multirow{2}{*}{0.563} \\
 &  & True & \textcolor{subblue}{Didier Marouani} was born in \textcolor{objorange}{Monaco}. & 4 & 0.0839 &  \\
\addlinespace[2pt] \midrule \addlinespace[2pt]
\multirow{2}{*}{P1303} & \multirow{2}{*}{314} & False & \textcolor{subblue}{Rosamund Pike} is a \textcolor{objorange}{viola} player. & 1 & 0.3485 & \multirow{2}{*}{0.522} \\
 &  & True & \textcolor{subblue}{Rosamund Pike} is a \textcolor{objorange}{cello} player. & 2 & 0.3046 &  \\
\addlinespace[2pt] \midrule \addlinespace[2pt]
\multirow{2}{*}{P206} & \multirow{2}{*}{198} & False & \textcolor{subblue}{Biscoe Islands} is situated on the shores of the \textcolor{objorange}{Indian Ocean}. & 1 & 0.7662 & \multirow{2}{*}{0.3375} \\
 &  & True & \textcolor{subblue}{Biscoe Islands} is situated on the shores of the \textcolor{objorange}{Southern Ocean}. & 2 & 0.0588 &  \\
\addlinespace[2pt] \midrule \addlinespace[2pt]
\multirow{2}{*}{P178} & \multirow{2}{*}{171} & False & \textcolor{subblue}{iOS} is developed by \textcolor{objorange}{Google}. & 1 & 0.4219 & \multirow{2}{*}{0.3706} \\
 &  & True & \textcolor{subblue}{iOS} is developed by \textcolor{objorange}{Apple Inc.}. & 2 & 0.3389 &  \\
\addlinespace[2pt] \midrule \addlinespace[2pt]
\multirow{2}{*}{P19} & \multirow{2}{*}{9} & False & \textcolor{subblue}{Zahir Khan} was born in \textcolor{objorange}{Pakistan}. & 1 & 0.6817 & \multirow{2}{*}{0.2711} \\
 &  & True & \textcolor{subblue}{Zahir Khan} was born in \textcolor{objorange}{Afghanistan}. & 2 & 0.2735 &  \\
\addlinespace[2pt] \midrule \addlinespace[2pt]
\multirow{2}{*}{P175} & \multirow{2}{*}{152} & False & \textcolor{subblue}{Love You Live} was performed by \textcolor{objorange}{David Bowie}. & 1 & 0.388 & \multirow{2}{*}{0.6016} \\
 &  & True & \textcolor{subblue}{Love You Live} was performed by \textcolor{objorange}{The Rolling Stones}. & 2 & 0.2469 &  \\
\addlinespace[2pt] \midrule \addlinespace[2pt]
\multirow{2}{*}{P466} & \multirow{2}{*}{277} & False & The occupant of \textcolor{subblue}{El Sadar Stadium} is \textcolor{objorange}{Real Betis Balompié}. & 1 & 0.6635 & \multirow{2}{*}{0.3132} \\
 &  & True & The occupant of \textcolor{subblue}{El Sadar Stadium} is \textcolor{objorange}{Club Atlético Osasuna}. & 5 & 0.0266 &  \\
\addlinespace[2pt] \midrule \addlinespace[2pt]
\multirow{2}{*}{P7937} & \multirow{2}{*}{365} & False & The \textcolor{subblue}{Swan of Tuonela} is a form of \textcolor{objorange}{waltz}. & 1 & 0.385 & \multirow{2}{*}{0.452} \\
 &  & True & The \textcolor{subblue}{Swan of Tuonela} is a form of \textcolor{objorange}{symphonic poem}. & 2 & 0.3495 &  \\
\addlinespace[2pt] \midrule \addlinespace[2pt]
\multirow{2}{*}{P7959} & \multirow{2}{*}{371} & False & \textcolor{subblue}{Quin} is located in the historic county of \textcolor{objorange}{Cornwall}. & 1 & 0.2946 & \multirow{2}{*}{0.6588} \\
 &  & True & \textcolor{subblue}{Quin} is located in the historic county of \textcolor{objorange}{County Clare}. & 4 & 0.0771 &  \\
\addlinespace[2pt] \midrule \addlinespace[2pt]
\multirow{2}{*}{P509} & \multirow{2}{*}{279} & False & \textcolor{subblue}{Cardinal Richelieu} died from \textcolor{objorange}{peritonitis}. & 1 & 0.3046 & \multirow{2}{*}{0.6528} \\
 &  & True & \textcolor{subblue}{Cardinal Richelieu} died from \textcolor{objorange}{tuberculosis}. & 3 & 0.1668 &  \\
\addlinespace[2pt] \midrule \addlinespace[2pt]
\multirow{2}{*}{P4552} & \multirow{2}{*}{349} & False & \textcolor{subblue}{Elidir Fawr} is part of the \textcolor{objorange}{Massif Central}. & 1 & 0.22 & \multirow{2}{*}{0.699} \\
 &  & True & \textcolor{subblue}{Elidir Fawr} is part of the \textcolor{objorange}{Snowdonia}. & 2 & 0.2023 &  \\
\addlinespace[2pt] \midrule \addlinespace[2pt]
\multirow{2}{*}{P466} & \multirow{2}{*}{275} & False & \textcolor{subblue}{Allianz Stadium} is occupied by \textcolor{objorange}{Stanford University}. & 1 & 0.2156 & \multirow{2}{*}{0.5922} \\
 &  & True & \textcolor{subblue}{Allianz Stadium} is occupied by \textcolor{objorange}{Juventus F.C.}. & 4 & 0.1084 &  \\
\addlinespace[2pt] \midrule \addlinespace[2pt]
\multirow{2}{*}{P364} & \multirow{2}{*}{231} & False & \textcolor{subblue}{Shake It All About} was originally created in \textcolor{objorange}{French}. & 1 & 0.2289 & \multirow{2}{*}{0.6675} \\
 &  & True & \textcolor{subblue}{Shake It All About} was originally created in \textcolor{objorange}{Danish}. & 10 & 0.0152 &  \\
\addlinespace[2pt] \midrule \addlinespace[2pt]
\multirow{2}{*}{P1532} & \multirow{2}{*}{331} & False & \textcolor{subblue}{Ronaldo} represents \textcolor{objorange}{Portugal}. & 1 & 0.9522 & \multirow{2}{*}{0.0806} \\
 &  & True & \textcolor{subblue}{Ronaldo} represents \textcolor{objorange}{Brazil}. & 2 & 0.0329 &  \\
\addlinespace[2pt] \midrule \addlinespace[2pt]
\multirow{2}{*}{P108} & \multirow{2}{*}{101} & False & \textcolor{subblue}{Mary Lou Williams} works for \textcolor{objorange}{Google}. & 1 & 0.2291 & \multirow{2}{*}{0.6821} \\
 &  & True & \textcolor{subblue}{Mary Lou Williams} works for \textcolor{objorange}{Duke University}. & 6 & 0.07 &  \\
\addlinespace[2pt] \midrule \addlinespace[2pt]
\multirow{2}{*}{P177} & \multirow{2}{*}{162} & False & \textcolor{subblue}{The Alfred H. Smith Memorial Bridge} crosses the \textcolor{objorange}{Connecticut River}. & 1 & 0.2193 & \multirow{2}{*}{0.7419} \\
 &  & True & \textcolor{subblue}{The Alfred H. Smith Memorial Bridge} crosses the \textcolor{objorange}{Hudson River}. & 3 & 0.1514 &  \\
\addlinespace[2pt] \midrule \addlinespace[2pt]
\multirow{2}{*}{P171} & \multirow{2}{*}{144} & False & The parent taxon of \textcolor{subblue}{Bifora} is \textcolor{objorange}{Asteraceae}. & 1 & 0.6527 & \multirow{2}{*}{0.3646} \\
 &  & True & The parent taxon of \textcolor{subblue}{Bifora}  is \textcolor{objorange}{Apiaceae}. & 2 & 0.1884 &  \\
\addlinespace[2pt] \midrule \addlinespace[2pt]
\multirow{2}{*}{P364} & \multirow{2}{*}{230} & False & The original language of \textcolor{subblue}{Padmaavat} is \textcolor{objorange}{Malayalam}. & 1 & 0.4069 & \multirow{2}{*}{0.5318} \\
 &  & True & The original language of \textcolor{subblue}{Padmaavat} is \textcolor{objorange}{Hindi}. & 2 & 0.2162 &  \\
\addlinespace[2pt] \midrule \addlinespace[2pt]
\multirow{2}{*}{P427} & \multirow{2}{*}{257} & False & The taxonomic type of \textcolor{subblue}{Rhizocarpon} is \textcolor{objorange}{Claviceps purpurea}. & 1 & 0.6416 & \multirow{2}{*}{0.3663} \\
 &  & True & The taxonomic type of \textcolor{subblue}{Rhizocarpon} is \textcolor{objorange}{Map lichen}. & 10 & 0.0071 &  \\
\addlinespace[2pt] \midrule \addlinespace[2pt]
\multirow{2}{*}{P291} & \multirow{2}{*}{214} & False & \textcolor{subblue}{Follow the Reaper} got published in \textcolor{objorange}{Japan}. & 1 & 0.1692 & \multirow{2}{*}{0.8415} \\
 &  & True & \textcolor{subblue}{Follow the Reaper} got published in \textcolor{objorange}{Finland}. & 11 & 0.0206 &  \\
\addlinespace[2pt] \midrule \addlinespace[2pt]
\multirow{2}{*}{P69} & \multirow{2}{*}{69} & False & \textcolor{subblue}{William Thomas Blanford} was educated at the \textcolor{objorange}{University of Oxford}. & 1 & 0.6478 & \multirow{2}{*}{0.2856} \\
 &  & True & \textcolor{subblue}{William Thomas Blanford} was educated at the \textcolor{objorange}{Imperial College London}. & 12 & 0.0003 &  \\
\addlinespace[2pt] \midrule \addlinespace[2pt]
\multirow{2}{*}{P449} & \multirow{2}{*}{262} & False & \textcolor{subblue}{The Son} was originally broadcasted by \textcolor{objorange}{CBS}. & 1 & 0.2422 & \multirow{2}{*}{0.736} \\
 &  & True & \textcolor{subblue}{The Son} was originally broadcasted by \textcolor{objorange}{AMC}. & 3 & 0.1433 &  \\
\addlinespace[2pt] \midrule \addlinespace[2pt]
\multirow{2}{*}{P206} & \multirow{2}{*}{200} & False & \textcolor{subblue}{Lipari} is situated on the shores of the \textcolor{objorange}{Ionian Sea}. & 1 & 0.402 & \multirow{2}{*}{0.392} \\
 &  & True & \textcolor{subblue}{Lipari} is situated on the shores of the \textcolor{objorange}{Tyrrhenian Sea}. & 2 & 0.3725 &  \\
\addlinespace[2pt] \midrule \addlinespace[2pt]
\multirow{2}{*}{P7937} & \multirow{2}{*}{363} & False & \textcolor{subblue}{Ombra mai fu} is a form of \textcolor{objorange}{opera}. & 1 & 0.4891 & \multirow{2}{*}{0.3061} \\
 &  & True & \textcolor{subblue}{Ombra mai fu} is a form of \textcolor{objorange}{aria}. & 2 & 0.4493 &  \\
\bottomrule
\end{tabular}
}
\caption{A random selection of errors originating from \texttt{Llama 2 13B} model spanning a diverse sample of relations. Each instance is constructed with a subject (highlighted in \textcolor{subblue}{blue}), a template (in plain text) and an object (highlighted in \textcolor{objorange}{orange}). Probability values are obtained by applying the softmax function to the log-likelihood scores of all potential answers. Uncertainty is measured by the similarity to a uniform distribution (see Section \ref{sec:predictions-confidence}). The lower the uncertainty value, the higher the model's confidence in its prediction.}
\label{tab:error-analysis}
\end{table*}

\begin{figure*}[t!]
    \centering

    \includegraphics[height=0.95\textheight]{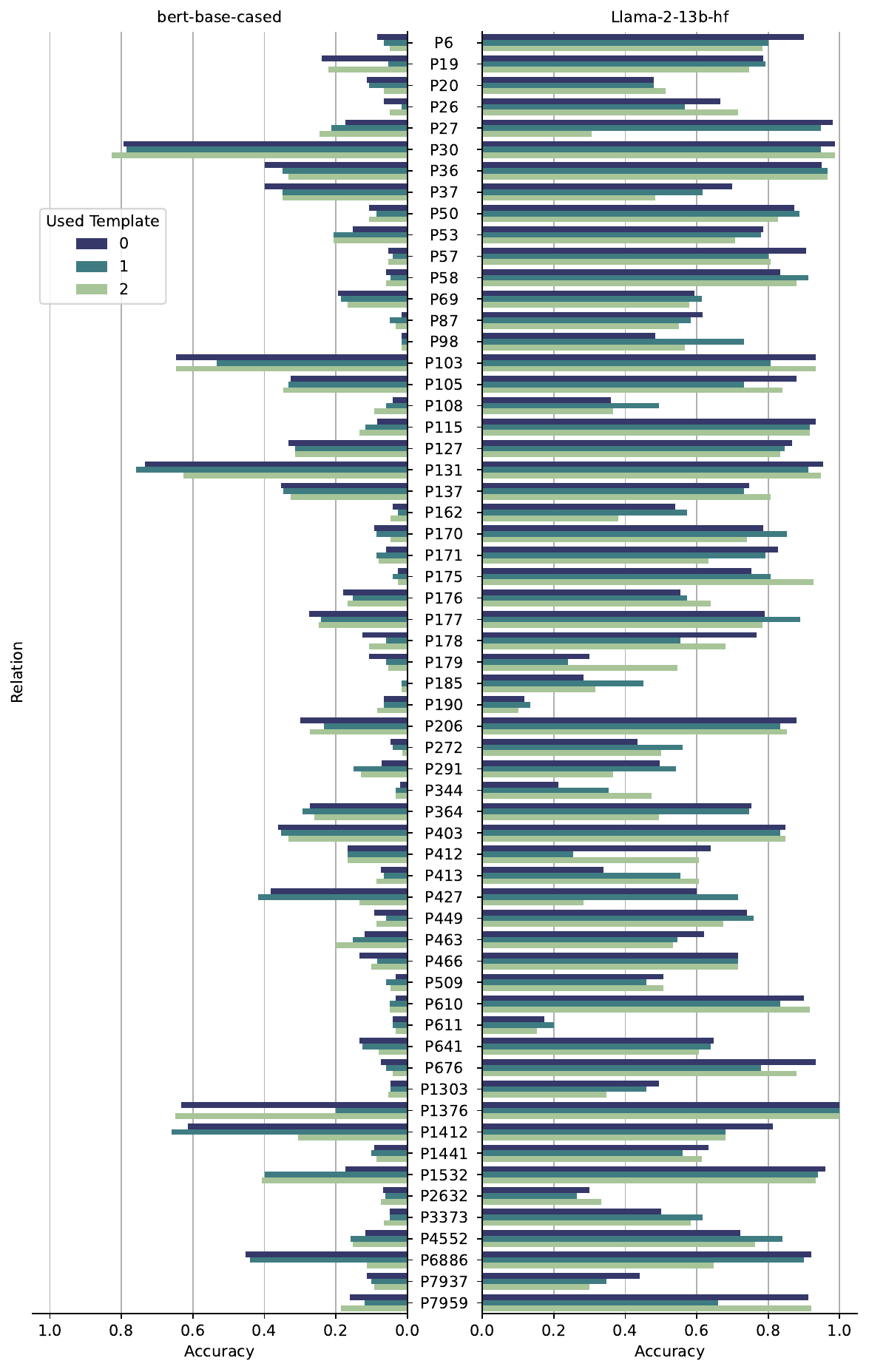}

    \caption{Accuracy of two models on each of the BEAR relations.}
    \label{fig:accuracy_per_relation}
\end{figure*}

\end{document}